\documentclass[11pt]{article}

\usepackage[preprint]{acl}

\usepackage{amsmath}
\usepackage{tabularx}
\usepackage{microtype}
\usepackage{booktabs}
\usepackage{enumitem}
\usepackage{multirow}
\usepackage[edges]{forest}
\usepackage{tikz-dependency}
\usepackage{fontawesome5}
\usepackage{placeins} 
\usepackage{arydshln} 

\usepackage{inconsolata}
\usepackage{newtxtext}
\usepackage{newtxmath}
\usepackage[T1]{fontenc}
\usepackage[utf8]{inputenc}
\usepackage{latexsym}

\newcommand{\phen}[1]{\textsc{#1}}

\title{Where’s the Indirect Evidence? Investigating Grammatical Generalization with Influence Functions}

\author{
    William Dai$^{1}$ \hspace{1em} Zhenghao Herbert Zhou$^{2}$ \hspace{1em} Maya Viswanathan$^{1}$ \hspace{1em} Robert Frank$^{2,3}$\\
    $^{1}$Department of Computer Science, Yale University\\
    $^{2}$Department of Linguistics, Yale University\\ 
    $^{3}$Wu Tsai Institute, Yale University\\
    \texttt{\{william.dai, herbert.zhou, maya.viswanathan,}\\
    \texttt{ robert.frank\}@yale.edu}
}
\begin{document}

\maketitle
\begin{abstract}
Language models generalize beyond their training corpus to learn linguistic phenomena \citep{patil2024filtered,misra2024language}, showing that there exists indirect evidence driving learning. Prior work has inferred that indirect evidence exists without identifying its character or considered specific hypotheses about what it might be. We explore an alternative bottom-up method for identifying indirect evidence. We use influence functions \citep{koh2017understanding} to measure how individual training sentences affect model predictions in 15 grammatical phenomena from BLiMP \citep{warstadt2020blimp}. We filter out a phenomenon's direct evidence using hand-constructed rules and take the 60,000 most influential remaining sentences as indirect evidence. This indirect evidence substantially affects learning: retraining without it lowers held-out accuracy more than removing a matched random group for every phenomenon across five models (by a median of $12$ points). We characterize the indirect evidence by its similarity to our direct evidence filters: it (partially) meets the filters' structural conditions at a median of $4.7$ times the rate of a matched random group. Finally, clustering phenomena by indirect-evidence overlap recovers three of the five BLiMP categories that include multiple phenomena, suggesting that this method can reveal linguistic relationships.
\end{abstract}

\section{Introduction}

\begin{figure*}[t]
    \centering
    \includegraphics[width=\textwidth]{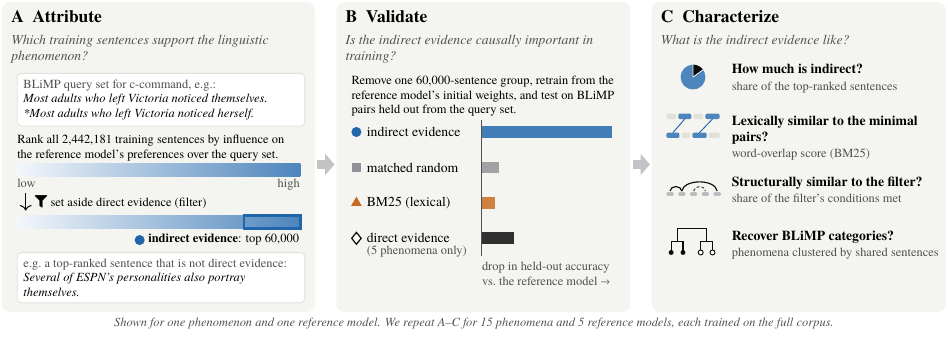}
    \caption{Schematic overview of our methodology, shown for one phenomenon (\phen{c-command}) and one reference model. (A) We rank every sentence in the training corpus using influence functions with an attribution objective capturing the model's grammatical preferences in the phenomenon's BLiMP minimal pairs. We filter out the phenomenon's direct evidence and take the 60,000 most influential remaining sentences as indirect evidence. (B) We remove the indirect evidence or an equally sized control group, retrain without the group, and then evaluate on held-out minimal pairs to validate our methodology. (C) We characterize the indirect evidence.}
    \label{fig:pipeline}
\end{figure*}

Language models are capable of learning grammatical phenomena that are rare or even absent in the training data \citep{patil2024filtered,misra2024language}.\footnote{However, other works find limits on how language models can generalize. For instance, \citet{oba2024can} find weak generalization from linguistic evidence, and generalization fails for some constructions \citep{lan2026large,howitt2024generalizations,zhou-etal-2026-exactly}.} This fact suggests that they generalize to the unattested forms using \emph{indirect evidence}: sentences which do not directly contain a phenomenon, but which have properties that are informative about it.

Unlike the input a child receives, the training data of a language model is fully observed and controllable. As a result, the study of language models offers an opportunity to investigate what indirect evidence can drive the learning of a range of phenomena \citep{warstadt2022artificial}.
\citet{patil2024filtered} individually remove direct evidence for a range of phenomena and show that learning is still possible, establishing that indirect evidence must exist but not what such evidence consists of. \citet{misra2024language} usefully explore the role of particular sources of indirect evidence in the case of the AANN construction. However, their approach can only confirm or disprove specific prior hypotheses about the nature of indirect evidence. It would be useful to understand the nature of indirect evidence in a broad range of cases, and in the absence of specific hypotheses about what data might be relevant for generalization. We would like to be able to determine empirically what features or cues are most relevant for learning a phenomenon.

In order to do this, we propose a bottom-up procedure for identifying indirect evidence  (Figure~\ref{fig:pipeline}). We use influence functions \citep{koh2017understanding} to evaluate, for a range of grammatical phenomena, how much each training sentence supports a model's preference for the grammatical sentences in BLiMP minimal pairs \citep{warstadt2020blimp}.  We then filter out sentences that constitute direct evidence using hand-constructed rules, and take the remaining sentences that are judged most influential as the relevant indirect evidence. We then test the significance of this evidence for learning each phenomenon by retraining without it. Finally, we characterize the indirect evidence by its lexical and structural similarity to each phenomenon.

This paper makes the following contributions:\footnote{Our code will be made publicly available upon acceptance.}
\begin{itemize}[leftmargin=*, itemsep=1pt, topsep=3pt, parsep=0pt]
\item We give an operational \textbf{bottom-up definition of indirect evidence} for syntactic capability, and apply it to 15 phenomena on five reference models.
\item We \textbf{validate the method through retraining}: in all 15 phenomena and all five reference models, removing a phenomenon's 60,000 indirect evidence sentences lowers held-out accuracy more than removing a matched random group.
\item We \textbf{characterize the indirect evidence}. Most of the top 60,000 sentences in the full-corpus influence ranking are indirect evidence. We find that while lexical similarity plays a modest role in establishing a sentence as indirect evidence, structural similarity plays a larger role. 
\item We \textbf{recover relationships between grammatical phenomena}:
the overlap in indirect evidence that we identify for grammatically similar phenomena is twice that of unrelated phenomena. Clustering of shared evidence largely recovers broad BLiMP categories, such as agreement and binding.
\end{itemize}

\section{Methodology}
\subsection{Influence functions}

Influence functions have been proposed as a way of measuring how a training example affects a model's behavior \citep{koh2017understanding}. Influence functions have been successfully exploited  in attributing LLM behavior in reasoning \citep{ruis2025procedural}, factual recall \citep{chang2025scalable}, and generalization more broadly \citep{grosse2023studying}. In this study, we apply influence functions to the study of what supports linguistic phenomena in syntax and semantics.

Influence functions estimate how a differentiable function $f(\theta)$ of the model parameters $\theta$ changes when a training sentence's weight changes. Let $\hat{\theta}$ be the model's fitted parameters, $\ell_{i}(\theta)$ be the language-modeling loss for training sentence $z_i$ and $f(\theta)$ the attribution objective. We define the gradients $g_{i}=\nabla_{\theta}\ell_{i}(\hat{\theta})$ and $q=\nabla_{\theta} f(\hat{\theta})$. We compute an estimate of the impact of the $i$th sentence on the attribution objective as follows:  
\[f(\hat{\theta}_{-i})-f(\hat{\theta}) \approx \frac{1}{N} q^T H^{-1} g_{i} \]
where $N$ is the number of sentences in the training corpus, $\hat{\theta}_{-i}$ denotes the model's parameters fitted without $z_{i}$, and $H$ is the Hessian of the average training loss at $\hat{\theta}$. For a strictly convex loss at its minimum, this cheaply approximates leave-one-out retraining. For neural networks, loss is not strictly convex and influence functions instead approximate the proximal Bregman response function \citep{bae2022if}. We can therefore interpret influence as how much the trained model's grammatical preferences respond to up-weighting the sentence, rather than a direct prediction of retraining without it.

This exact computation cannot be feasibly done for LLMs because of the difficulty of computing the inverse Hessian. Instead, scaling to LLMs requires approximations and other approaches, such as EK-FAC \citep{grosse2023studying}, linearization of the model with the neural tangent kernel \citep{park2023trak}, or random projection of the gradients \citep{lin2024rapidin}. 
We approximate the inverse Hessian term $H^{-1}$ with damped EK-FAC \citep{grosse2023studying}, computed with Kronfluence (Appendix~\ref{app:influence}). Then, letting $B$ be the resulting approximation to the inverse Hessian operator, the relative-influence score is the following:
\[r_{i}=\frac{q^TBg_{i}}{\sqrt{g_{i}^T Bg_{i}}}\]
where we apply a normalization term \citep{barshan2020relatif}; it normalizes by the training gradient's norm with the EK-FAC approximation. Normalization favors training sentences whose gradients align with the gradient of the attribution objective, and keeps sentences with large gradients from dominating.
\subsection{Attribution objective}
\label{sec:objective}
We now construct an attribution objective that captures a language model's grammatical preferences for linguistic phenomena. To evaluate a language model's grammatical preferences on a particular linguistic phenomenon, we use minimal pairs: two sentences that differ minimally but show a contrast in grammaticality because of adherence (or not) to some grammatical regularity. For example,  \textit{Most adults who left Victoria noticed themselves} and *\textit{...noticed herself} constitute a minimal pair for the phenomenon of \phen{c-command}. However, we cannot simply attribute on minimal pair accuracy because it is a binary metric which either has zero gradient or is discontinuous, making it unsuitable for influence functions computation. Instead, we need an attribution objective that is smooth.

Let $x^{+}$ and $x^{-}$ be the grammatical and ungrammatical members of a minimal pair. For a model with parameters $\theta$, we define the \textbf{margin} as follows:
\[m_{\theta}(x^{+}, x^{-})=s_{\theta}(x^{+})-s_{\theta}(x^{-})\]
where $s_{\theta}(x)$ is the log-probability score of $x$. A positive margin means the model (correctly) prefers the grammatical sentence in the pair. Then, using the margin, we define the \textbf{logistic margin loss}:
\[\ell_{\text{pair}}(\theta; x^{+}, x^{-}) = -\log \sigma(m_{\theta}(x^+,x^-))\]
This loss has the property that its gradient, $\nabla_{\theta} \ell_{\text{pair}} = -\sigma(-m_{\theta}) \nabla_{\theta} m_{\theta}$, weights each pair by $\sigma(-m_{\theta})$. This means that the weight is small for pairs that have high margin and thus are confidently predicted correctly (so those individual examples are less likely to be impactful), while it is larger for pairs that have low margin (so they are more likely to be affected). Hence, reducing this logistic margin loss targets weak grammatical preferences. 

The logistic margin loss can be extended from minimal pairs to entire phenomena. For each phenomenon, we have a query set of minimal pairs for influence function scoring; a separate held-out set of minimal pairs is used for evaluating the effects of removal (Section~\ref{subsec:splits}). Let $\mathcal{P}_{t}$ be the BLiMP paradigms (the minimal pair types) associated with phenomenon $t$ and $\mathcal{Q}_{p}$ be the query minimal pairs from paradigm $p$. The attribution objective then is:
\[\mathcal{L}_{t} = \frac{1}{|\mathcal{P}_{t}|} \sum_{p\in \mathcal{P}_{t}} \frac{1}{|\mathcal{Q}_{p}|} \sum_{(x^{+},x^{-})\in \mathcal{Q}_{p}} \ell_{\text{pair}}(\theta; x^{+},x^{-})\]

This phenomenon-level loss averages over the minimal pairs within each paradigm, then equally over paradigms so that paradigm has equal weighting even if they have a differing number of retained pairs.

\section{Experimental Setup}
\subsection{Model and corpus}
\paragraph{Corpus.} We train, attribute, and retrain using the English Wikipedia corpus of \citet{gulordava2018colorless}, also used by \citet{patil2024filtered}, which uses a 50k-word vocabulary with rare words being mapped to \texttt{<unk>}. Also following \citet{patil2024filtered}, we uniformly subsample 80\% of the 3,052,726 training sentences. We use this subsample for attribution, removal, and characterization.\footnote{Hereafter, we use the term ``full corpus" to refer to this 80\% subsample.}

\paragraph{Architecture.} Following \citet{patil2024filtered}, we explore OPT-style decoder-only transformers \citep{zhang2022opt}. These models have 8 layers, 8 attention heads, hidden and feed-forward dimension 768, a word-level tokenizer with a vocabulary of size 50,002, and ${\sim}$67M parameters. Models are all trained for 40 epochs. Appendix~\ref{app:compute} lists details of hardware and compute.

\paragraph{Reference models.} We train five reference models from scratch following the same architecture, training corpus, and hyperparameters as \citet{patil2024filtered} (see Appendix~\ref{app:models}). Our reference models are trained on the full corpus. We perform attribution on each model at the checkpoint with lowest validation loss. For each phenomenon, we identify supporting sentences that are outside of the direct evidence filter in models trained on the full corpus. 

\subsection{Phenomena and splits}
We study the fifteen \emph{phenomena} of \citet{patil2024filtered} (Table~\ref{tab:phenomena}, Appendix~\ref{app:phenomena}), each with one or more BLiMP paradigms that evaluate it. For filtering direct evidence, we use \citeauthor{patil2024filtered}'s twelve parse-based filters and three Negative Polarity Item (NPI) surface filters (Appendix~\ref{app:filters}). The \phen{regular/irregular plural}, \phen{determiner--noun}, and \phen{passive} filters test lexical generalization and remove a construction only with BLiMP's words \citep[\S3.2.2]{patil2024filtered}. BLiMP also groups its 67 paradigms into 12 broader classes, which we call \emph{categories}.   
\label{subsec:splits}
For each phenomenon, we randomly split its minimal pairs into a \emph{query} set and a \emph{held-out} set. We randomly draw $500$ query pairs (with an equal proportion from each paradigm testing the phenomenon) and hold out the rest of the pairs for evaluation.\footnote{The query set is smaller overall than the held-out set because the query set only has to estimate one phenomenon-level query gradient, whereas the held-out set has to detect possibly small intervention effects.} The minimal pairs in the query set are used to define the attribution objective (Section~\ref{sec:objective}). Meanwhile, the held-out pairs are used to measure the effect of a data intervention in Section~\ref{sec:validation}. Because evidence is selected using only the query set, its effect on the held-out minimal pairs' logistic margin loss does not follow from how influence is defined. Additionally, we exclude minimal pairs containing \texttt{<unk>} as well as 6 pairs whose two sentences are identical (Appendix~\ref{app:phenomena}). In total, this leaves 5,826 query pairs (with a range of 307 to 462 per phenomenon) and 18,416 held-out pairs.

\subsection{Control groups}
\label{sec:groups} 
For each phenomenon $t$ and reference model, we take the \textbf{indirect evidence} group to be the 60,000 most influential sentences which do not match $t$'s filter (about 2.5\% of the corpus).\footnote{Appendix~\ref{app:removal-size} compares the result if we consider the 20,000 most influential sentences.} We compare the effects of the indirect evidence training data to the following control groups (details are elaborated in Appendix~\ref{app:matching}):

\begin{itemize}[leftmargin=*, itemsep=1pt, topsep=3pt, parsep=0pt]
    \item \textbf{Matched random.} For each indirect evidence sentence, we draw a random sentence of the same length and similar mean log word frequency from the remaining non-direct sentences.
    \item \textbf{BM25.} The purpose of this control group is to compare our influence-based selection against lexical selection. We identify the top 60,000 sentences by BM25 score, a measure of lexical similarity \citep{robertson2009bm25}, against the vocabulary of $t$'s query pairs.  
    \item \textbf{Direct Evidence.} The sentences that match $t$'s filter. For characterization, we use all of them. For retraining, we remove a random sample of 60,000, the same size as the other groups. Only five phenomena have that many direct-evidence sentences (\phen{relational-noun distractor}, \phen{relative-clause distractor}, \phen{regular/irregular plural}, \phen{superlative}, and \phen{passive}), so retraining covers only those five.
\end{itemize}

We draw every group except direct evidence from the pool of sentences that do not match the direct evidence filter. The groups therefore differ only in how they are selected: by influence, at random, or by lexical similarity.

\section{Causal Validation}
\label{sec:validation}
To test whether the identified indirect evidence is causally relevant, we remove it from the training corpus and retrain from scratch (Panel B in Figure~\ref{fig:pipeline}). For each phenomenon and reference model, we remove one group, either indirect evidence or one of the controls (Section~\ref{sec:groups}), from the training corpus and then train a new model using the reference model's initialization, data order, and hyperparameters. We then evaluate on the held-out pairs at the same epoch as the reference model's best-validation checkpoint (we test alternative choices for this in Appendix~\ref{app:par:checkpoint}); every comparison is exactly paired between the control and a reference model with the same initialization. We expect model performance to significantly degrade when the indirect evidence is removed and to not degrade as much when the control groups are removed.

\begin{figure}[t]
    \centering
    \includegraphics[width=\columnwidth]{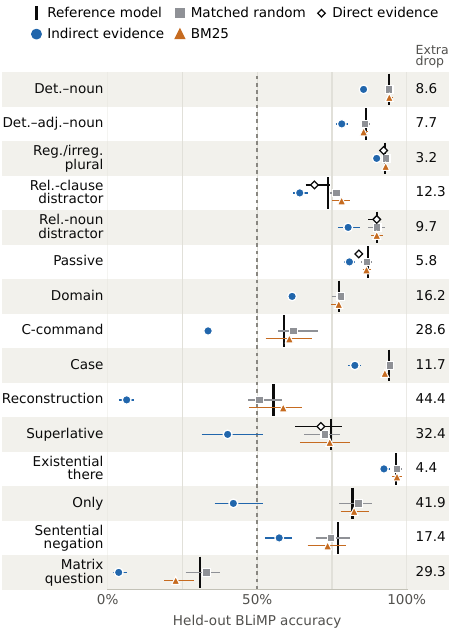}
    \caption{Accuracy on held-out BLiMP minimal pairs after removing each group. Extra drop is the additional drop in accuracy, averaged over the five initializations, when removing indirect evidence compared to matched random. Points show means over the five models; bars are the range. Ranges narrower than a marker are hidden behind it. Comparing against direct evidence (a random 60,000 of its sentences) was done only for the five phenomena with at least 60,000 direct-evidence sentences.}
    \label{fig:removal-accuracy}
\end{figure}

\paragraph{Outcome.} Figure~\ref{fig:removal-accuracy} shows the results of removing indirect evidence and each control group. Removing indirect evidence lowers held-out BLiMP accuracy more than removing a matched random group for all 15 phenomena in each of the five reference models. On held-out loss, an exact sign-flip test across the five reference models gives $p=1/32$, its smallest possible value (Appendix~\ref{app:validation}). Relative to the matched random group and averaged over the five initializations, the extra accuracy drop ranges from $3.2$ to $44.4$ percentage points, with a median of about $12$ points.\footnote{Percentage points are on an absolute scale and not relative.} Both groups are the same size and matched on sentence length and mean word frequency. Hence, the larger accuracy drop shows that influence functions select for more influential sentences than random. We also see larger accuracy drops for indirect evidence as compared to the BM25 control group and, perhaps surprisingly, the direct evidence group. The BM25 results demonstrate that lexical similarity alone does not imply causal importance in retraining. The direct evidence results could be explained by the fact that the removal of all direct evidence in \citet{patil2024filtered} generally results in modest effects, and we only remove a fraction of the direct evidence; this may also suggest direct evidence does not need to be highly influential.

The largest drops from removing indirect evidence generally come from phenomena that are not robustly learned by the reference models, such as \phen{reconstruction} (55\% held-out accuracy) and \phen{matrix question} (31\% held-out accuracy). We discuss this further in Section~\ref{disc:sensitivity}.

\begin{figure}[t]
    \centering
    \includegraphics[width=\columnwidth]{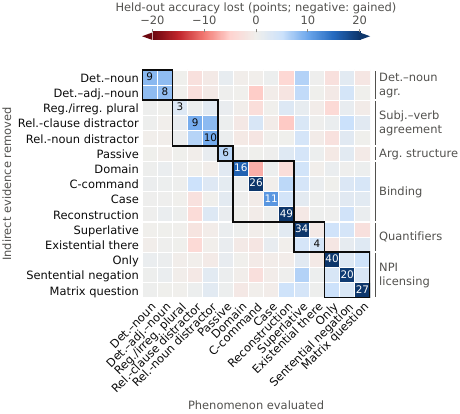}
    \caption{Held-out accuracy lost (points) after removing each phenomenon's indirect evidence (rows), evaluated on every phenomenon (columns), relative to the reference model and averaged over five initializations. Numbers give each phenomenon's own accuracy drop. Black boxes outline the BLiMP categories.}
    \label{fig:specificity}
\end{figure}

\paragraph{Same-category vs.\ different-category removal effects.} We want to understand the effect of a phenomenon's indirect evidence on other phenomena, and if this effect differs between same-category and different-category phenomena. 
Relative to the reference model and averaged over the five initializations, removing a phenomenon's own indirect evidence lowers its held-out accuracy by a median of $11.4$ points while removing another phenomenon's indirect evidence lowers accuracy by a median of $0.3$ points (Figure~\ref{fig:specificity}). Thus, the indirect evidence does not consist of sentences that are useful for every phenomenon. However, there does exist a causal relationship between phenomena in the same BLiMP category: same-category phenomena pairs have an average accuracy drop of $2.2$ points while different-category phenomena pairs have an average accuracy drop of $0.4$ points. Indirect evidence for a phenomenon is thus more important for phenomena in the same category than for phenomena in other categories, showing how indirect evidence may link together same-category phenomena.

\section{Characterizing Indirect Evidence}
Retraining helps to validate that our definition of indirect evidence yields data that supports the model's grammatical preferences for the relevant phenomenon. Unlike previous studies, the indirect evidence is not a priori characterized by prior hypotheses about what data might matter, so characterizing it is far from trivial. So, to further understand why the indirect evidence may aid in generalization, we ask how much of the top evidence is indirect and analyze the indirect evidence along three dimensions: lexical similarity, structural coverage, and overlap between phenomena. We also compare the indirect evidence to the control groups in Section~\ref{sec:groups}. These analyses illustrate what sentence-level evidence makes possible. To foreshadow the results, we find that the indirect evidence tends to share structure with its phenomenon and to be shared by related phenomena.

\begin{figure}[t]
    \centering
    \includegraphics[width=\columnwidth]{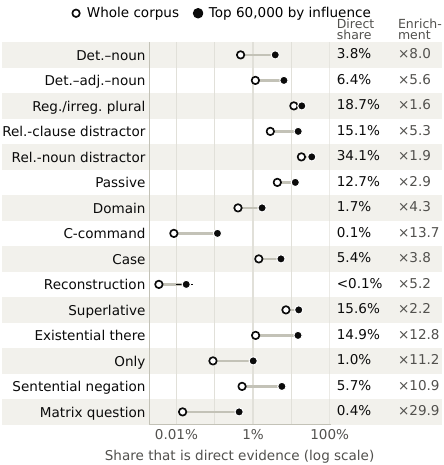}
    \caption{Direct evidence among the 60,000 most influential sentences compared to its proportion of the entire training corpus. Direct share is how much of the top 60,000 is direct evidence. Enrichment is direct share divided by how much of the corpus is direct evidence.}
    \label{fig:direct-evidence-share}
\end{figure}

\paragraph{How much of the evidence is direct?} To measure the share of indirect evidence, we rank every training sentence by its influence and compute the proportion of each phenomenon's 60,000 most influential sentences that are direct evidence as identified by its filter (Figure~\ref{fig:direct-evidence-share}). Influence favors direct evidence: it is enriched in the top 60,000 sentences compared to its share in the corpus. Still, direct evidence is at most $34.1\%$ for any phenomenon's top 60,000 (\phen{relational-noun distractor}) and under $19\%$ for the other $14$ phenomena. Table~\ref{tab:direct-share-by-k} in Appendix~\ref{app:top-k} gives other cutoffs. Most of each phenomenon's top 60,000 is therefore indirect evidence.

\begin{figure}[!t]
    \centering
    \includegraphics[width=\columnwidth]{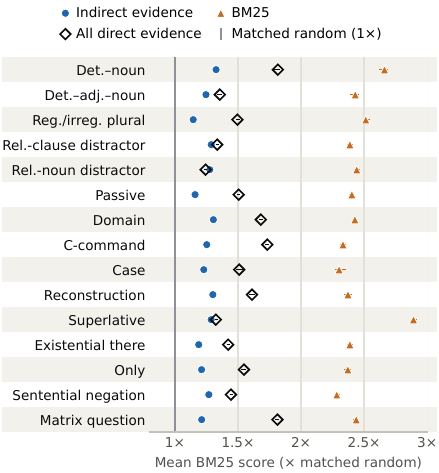}
    \caption{Lexical similarity of the indirect evidence, direct evidence, and the BM25 group, relative to the matched random group's lexical similarity. Points show means over the five models; bars are the range.}
    \label{fig:lexical-similarity}
\end{figure}

\paragraph{Is the indirect evidence lexically similar to the query pairs?} A deflationary hypothesis for why indirect evidence might be influential is that it shares lexical items with the query pairs. To test this hypothesis, we measure a group's \emph{lexical similarity} as the mean BM25 score of the sentences in that group against the vocabulary of the query pairs. The indirect evidence is slightly more lexically similar to the query pairs than the matched random group ($1.15$--$1.33\times$ its mean BM25 score), but in every phenomenon it has lexical similarity about half that of the BM25 group. Indirect evidence is also less lexically similar than the direct evidence for 14 of the 15 phenomena (Figure~\ref{fig:lexical-similarity}). Lexical similarity as measured by BM25 does not explain why the indirect evidence is influential.

\begin{figure}[t]
    \centering
    \includegraphics[width=\columnwidth]{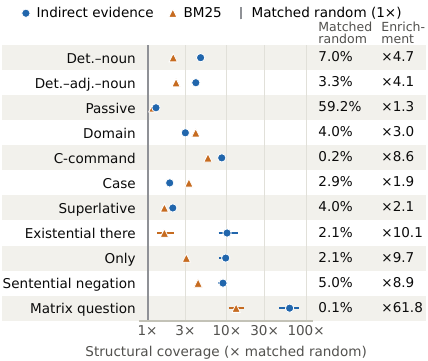}
    \caption{Structural coverage of the indirect evidence and the BM25 group relative to the matched random group (log scale), for the 11 phenomena whose coverage varies among non-direct sentences; Appendix~\ref{app:filters} reports the other four. Points are means over five models and bars are ranges; the right columns give the matched random group's mean coverage and the indirect evidence's enrichment.}
    \label{fig:structural-coverage}
\end{figure}

\paragraph{Is the indirect evidence structurally similar to the phenomenon?} Another hypothesis for why indirect evidence impacts learning is that indirect evidence shares syntactic structure with the phenomenon. While indirect evidence does not match all of the conditions in a phenomenon's filter, it may match some of the filter's structural conditions. We define a sentence's \emph{structural coverage} as the fraction of the filter's structural conditions it meets. To aid in understanding this definition, Table~\ref{tab:structural-similarity-examples} illustrates three examples of indirect evidence and the structural and lexical conditions they do or do not meet (Table~\ref{tab:filter-conditions} in Appendix~\ref{app:filters} enumerates all filters' conditions). In all 11 phenomena, the indirect evidence has higher structural coverage than the matched random group, ranging from $1.3\times$ to $62\times$ with a median of $4.7\times$ (Figure~\ref{fig:structural-coverage}). The indirect evidence's structural coverage is also higher than the BM25 group's in 9 (all but \phen{binding domain} and \phen{binding case}). Where coverage varies, indirect evidence is enriched for the structure its phenomenon's filter looks for, even though it is not direct evidence.

\begin{table}[t]
\centering
\footnotesize
\setlength{\dashlinedash}{2pt}
\setlength{\dashlinegap}{2pt}
\begin{tabularx}{\columnwidth}{@{}l@{\hspace{5pt}}>{\raggedright\arraybackslash}X@{}}
\toprule
\multicolumn{2}{@{}l}{\textbf{Binding C-command}} \\
Minimal pair & \textit{A lot of patients who can sell some couch didn't investigate \textbf{themselves} / *\textbf{itself}.} \\
\addlinespace[2pt]\hdashline\addlinespace[2pt]
\begin{tabular}[t]{@{}l@{}}Indirect\\Evidence\end{tabular} & \textit{Several of ESPN's personalities also portray \textbf{themselves}.} \\
\addlinespace[2pt]\hdashline\addlinespace[2pt]
Structural & \faCheck\ the reflexive's verb has a subject \\
 & \faTimes\ relative clause inside the subject \\
Lexical & \faCheck\ reflexive (\textit{themselves}) \\
\midrule
\multicolumn{2}{@{}l}{\textbf{Binding domain}} \\
Minimal pair & \textit{Carla had explained that Samuel has discussed \textbf{her} / *\textbf{herself}.} \\
\addlinespace[2pt]\hdashline\addlinespace[2pt]
\begin{tabular}[t]{@{}l@{}}Indirect\\Evidence\end{tabular} & \textit{Louise assumes that Warren is behind it and confronts \textbf{him}.} \\
\addlinespace[2pt]\hdashline\addlinespace[2pt]
Structural & \faCheck\ \textit{that} complement clause \\
 & \faTimes\ clause's head has an object (\textit{him} attaches to \textit{confronts}) \\
Lexical & \faTimes\ clause head's object is a pronoun (\textit{it} heads the clause; \textit{him} attaches to \textit{confronts})\\
\midrule
\multicolumn{2}{@{}l}{\textbf{Determiner--noun agreement}} \\
Minimal pair & \textit{Raymond is selling this \textbf{sketch} / *\textbf{sketches}.} \\
\addlinespace[2pt]\hdashline\addlinespace[2pt]
\begin{tabular}[t]{@{}l@{}}Indirect\\Evidence\end{tabular} & \textit{It is not positioned to compete with \textbf{those products}.} \\
\addlinespace[2pt]\hdashline\addlinespace[2pt]
Structural & \faCheck\ demonstrative determiner (\textit{those}) \\
Lexical & \faTimes\ noun on the BLiMP list (\textit{products} is not) \\
\bottomrule
\end{tabularx}
\caption{Examples of indirect evidence, with a minimal pair for each phenomenon
(* marks the ungrammatical sentence). Each is among the 60,000 most influential non-direct sentences of all five
reference models. \faCheck/\faTimes{} indicate the filter conditions it satisfies or misses. Structural conditions count
toward structural coverage, while lexical (word-list) conditions do not.}
\label{tab:structural-similarity-examples}
\end{table}

\begin{figure}[!h]
    \centering
    \includegraphics[width=\columnwidth]{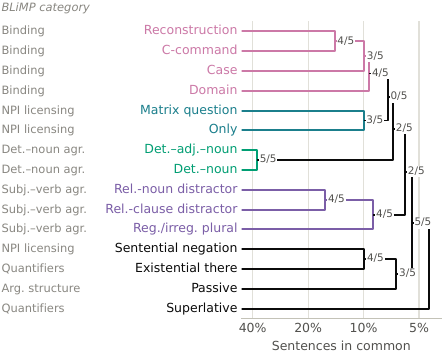}
    \caption{Average-linkage clustering of the 15 phenomena by shared indirect evidence. Distance between two phenomena is one minus overlap. The horizontal axis (log scale) gives the average overlap where two branches join. Colors are added after clustering and mark branches whose phenomena all belong to one BLiMP category. The ``x/5'' numbers at joins show how many of the five reference models produce the same cluster when they are clustered separately.}
    \label{fig:category_dendrogram}
\end{figure}

\paragraph{Do related phenomena share indirect evidence?} If indirect evidence carries structure that is useful beyond a single phenomenon, phenomena in the same category should share it \citep{ misra2024language}. For a pair of phenomena, we define \emph{overlap} as the size of the intersection between the two phenomena's indirect evidence set divided by the size of indirect evidence set (which is always 60,000), averaged over the five reference models. On average, phenomena in the same category have an overlap of $11.7\%$ against $5.8\%$ for phenomena in different categories. The exception is the two quantifier phenomena, which have an overlap of $5.0\%$, less than the $5.8\%$ average for different-category pairs. We also see that \phen{c-command} and \phen{relative-clause distractor} have an overlap of $14.1\%$, likely because the sentences from BLiMP that evaluate both phenomena involve relative clauses (Appendix~\ref{app:sharing} shows results for every pair). We also perform average-linkage clustering on the phenomena by their overlap (Figure~\ref{fig:category_dendrogram}). The clustering puts three of the five BLiMP categories with more than one phenomenon on their own branches: determiner--noun agreement, subject--verb agreement, and binding. However, the clustering also links together some different-category phenomena: for instance, \phen{sentential negation} and \phen{existential there}. This may be because both phenomena involve negation. Overall, this suggests we can recover linguistic categories using the indirect evidence. 

\FloatBarrier 
\section{Discussion and Conclusion}
\paragraph{Recovering category relationships.} The retraining results in \citet{patil2024filtered} demonstrate the existence of indirect evidence, but do not establish possible relationships between different phenomena in the same category. Indeed, in their results for transformer models, removing one phenomenon's direct evidence does not, on average, lower accuracy on same-category phenomenon more than on different-category phenomenon. Thus, removing direct evidence may not be enough to reveal relationships between same-category phenomena. This may be because direct evidence makes up a modest portion of what the model uses to learn, a median of $5.7\%$ of a phenomenon's 60,000 most influential sentences and at most $34.1\%$ (Figure~\ref{fig:direct-evidence-share}). In contrast, investigating identified indirect evidence does reveal same-category relationships (Figure~\ref{fig:category_dendrogram}, Figure~\ref{fig:specificity}, Figure~\ref{fig:category-membership} in Appendix~\ref{app:sharing}). These same-category phenomena could be linked by structurally enriched sentences that are not direct evidence for either phenomenon.

\paragraph{Removal sensitivity as a diagnostic.}\label{disc:sensitivity} The magnitude of the accuracy drop when a fixed amount of indirect evidence is removed indicates how concentrated the support is for the phenomenon. Across the 15 phenomena, the extra accuracy drop is higher when the reference model has lower competence on the phenomenon (Spearman $\rho=-0.78$ between reference model accuracy and extra accuracy drop). This indicates that these barely learned phenomena rely on a small set of sentences or on fragile surface regularities. When accuracy drop and BLiMP performance diverge, such as when the models score $82\%$ on \phen{only} but removal of indirect evidence leads to a drop of $42$ points relative to matched random, this may suggest that the phenomenon is actually fragile despite good BLiMP performance. Influence-guided removal or analysis of influence score concentration could mark phenomena that appear to be robustly learned, but are actually fragile.

\paragraph{A testbed for training data attribution.} It may be hard to judge whether an attribution method is working because the theory for which training examples should matter for a capability is often not developed or independent. In fact-tracing benchmarks measuring factual knowledge in a language model, the relevance of a training example could be defined through fact-entailment \citep{akyurek2022towards} or causally, for instance, through tail-patching or retraining. These two accounts of relevance may disagree, leading to high influence examples that do not entail a fact \citep{chang2025scalable}.  When an influential training example does not entail a fact, due to the lack of a developed theory, it is unclear whether it is meaningful or noise from a poor attribution method; even if causally influential, it is hard to interpret. Meanwhile, linguistics is a developed source of theories for generalization, and this could help us interpret influential training examples. Thus, we suggest syntactic generalization could be a testbed for attribution methods. Influential training examples that do not directly contain a linguistic phenomenon could be interpreted using descriptive terms from linguistic theory. This provides another check that our attribution method has reasonable results. For example, recovering BLiMP categories in Figure~\ref{fig:category_dendrogram} helps show that data attribution is finding sensible structure.

\paragraph{Open questions.} Our study leaves several questions open. First, our structural coverage metric is deliberately generic, so it does not give any account of which structures matter for a given phenomenon. Taking the relations in a sentence's dependency parse as features would show which structures are over-represented in the indirect evidence. Such an analysis could indicate whether the patterns of generalization adhere to the landscape of linguistic theory, for instance whether binding phenomena share indirect evidence containing reflexives. Second, pieces of evidence might play competing roles: sentences supporting one phenomenon might interfere with others. This could be revealed by measuring negative influence on other phenomena. Third, performing this study on developmentally plausible corpora could show whether child-directed speech provides a different profile of indirect evidence that enables rapid generalization.

\section*{Limitations}
Our study inherits the weaknesses of other studies applying influence functions to LLMs. Influence functions approximate the proximal Bregman response rather than leave-one-out retraining \citep{bae2022if}. The validation experiment does not mean the influence scores predict the magnitude of the retraining's impact. The Hessian must be approximated for influence functions to scale to LLMs. Influence functions may not capture redundancy in a group of training sentences where the removal of the entire group is needed for a measurable effect. Our methodology also requires fixing the parameter $k$ for determining what counts as an influential sentence, and we do not perform a systematic study over how changing $k$ may change our results.

Our language model (67M parameters) and training corpus are relatively small. To validate our methodology, we trained $270$ models with different groups of data removed. This retraining-based validation would be infeasible on a larger scale with our compute budget. However, \citet{khaddaj2025smalltolarge} find that training data tends to affect small and large models in correlated ways, so the method may remain valid at larger scale. On the other hand, \citet{grosse2023studying} find that the influential data for larger models tend to be less lexically similar to the queries. We may expect the lexical and structural profiles to change if we trained larger models. 

The scope of our exploration is also limited. We study only English, investigate only the 15 BLiMP phenomena that \citet{patil2024filtered} filtered, train only on one Wikipedia corpus, and investigate one architecture. BLiMP's minimal pairs are generated from templates with a restricted vocabulary, so held-out accuracy only measures performance on these templates rather than each phenomenon in general. For phenomena the models have not robustly learned, removing indirect evidence may just shift preferences related to the BLiMP minimal pairs templates' frames.

Influence functions attribution is relatively automatic, but our definition of indirect evidence still requires hand-written filters for direct evidence. These filters are imperfect and may leak direct examples of the phenomena; our study's filters inherit the same possible flaws as \citet{patil2024filtered}'s filters. We did not audit a sample of the indirect evidence by hand, so the extent of the filter leakage is unknown, though we should expect relatively little leakage; \citet{patil2024filtered} intentionally favored stronger filters which filter out more data, though at the cost of false positives. Our structural coverage metric also inherits the filters' design and conditions, and therefore is affected by particular design choices in its logic.

\section*{Broader Impact and Ethical Considerations}
This study had no human participants. The validation of attribution via retraining was computationally expensive, and took about 5,700 GPU-hours. Thus, this type of validation through retraining can be environmentally damaging; computationally lighter methods may reduce this harm. We do not foresee any risks arising from this research in particular beyond the general risks associated with machine learning research or data attribution research; moreover, the specific use of data attribution here is removed from potential harms such as data poisoning or adversarial manipulation of a dataset.
\bibliography{arxiv_main.bib}

\appendix
\renewcommand{\dbltopfraction}{0.9}
\renewcommand{\dblfloatpagefraction}{0.8}
\setcounter{dbltopnumber}{2}

\section{Reference Models}
\label{app:models}
We train five reference models with the same architecture, corpus, and hyperparameters as \citet{patil2024filtered} (Table~\ref{tab:training-settings}). We reconstruct \citeposs{patil2024filtered} training sample by rerunning their sampling code (seed 42, dropping sentences longer than 512 words). The resulting subsampled corpus has the same token count as the corpus in their released results. Our validation retrains each validation model from the same initialization as its corresponding reference model (\S\ref{sec:validation}). Because initializations are not fixed in \citeposs{patil2024filtered} code, we do not replicate their models exactly.

Moreover, there is a slight difference in how data order is handled. \citet{patil2024filtered} read the corpus in shards with parallel loaders. In our training, we shuffle all sentences once per epoch. We use the same shuffled data order and the same seed for training randomness for our five reference models; our reference models differ only in their initialization. This is to keep data order fixed for comparisons.

\begin{table}[h]
\centering
\small
\begin{tabularx}{\columnwidth}{@{}l>{\raggedright\arraybackslash}X@{}}
\toprule
Setting & Value \\
\midrule
Architecture & OPT decoder: 8 layers, 8 heads, hidden and feed-forward size 768, 512 positions \\
Parameters & 67,170,816 \\
Vocabulary & 50,002 tokens (includes padding token, \texttt{<unk>}, \texttt{<eos>}) \\
Training data & 2,442,181 sentences \\
Optimizer & AdamW ($\beta=(0.9, 0.999)$, $\epsilon=10^{-8}$), weight decay 0 \\
Learning rate & $5\times10^{-5}$, linear decay to 0, no warmup \\
Batch size, epochs & 32 sentences, 40 epochs \\
Gradient clipping & 1.0 \\
Dropout & 0.1 (none on attention) \\
Precision & fp16 mixed precision \\
Initializations & Seeds 2000--2004, one shared data order \\
Checkpoint & Each reference model's lowest-loss epoch on the validation split: 14, 14, 14, 15, 17 
\\
Hardware & One NVIDIA H100 per run, 13--27 hours (median 23) \\
\bottomrule
\end{tabularx}
\caption{Training settings of the reference and validation models.}
\label{tab:training-settings}
\end{table}

For every trained model used for validation, we start from its corresponding reference model's initialization. Then, the validation model sees the data in the same order as the corresponding reference model, except without the removed sentences. When reporting held-out accuracy or loss, we evaluate the validation model at its reference model's selected epoch rather than its own best epoch. This is so the validation model has the same number of passes over the data between evaluations in order to make comparison between models fair.

We provide some more details and elaboration on our metrics. For a sentence $x$, $s_\theta(x)$ is the sum of its token log-probabilities from the second word through an appended end-of-sentence token, without length normalization. We consider the language model to be correct on a minimal pair when $s_\theta(x^+) > s_\theta(x^-)$, where $x^{+}$ is the grammatical sentence and $x^{-}$ is the ungrammatical sentence. Our held-out accuracy and loss (the logistic margin loss of \S\ref{sec:objective}) metrics are calculated by averaging pairs within each paradigm, then averaging over paradigms with equal weight.

We also demonstrate comparable performance with the published models in \citet{patil2024filtered}. Table~\ref{tab:local-vs-published} compares our reference models with \citet{patil2024filtered}'s published full-corpus models. Both our reference models and the \citet{patil2024filtered} models are scored with our evaluation code on our held-out minimal pairs. The difference in BLiMP performance is generally minimal between our models and \citeposs{patil2024filtered} models. There is a median difference of $0.5$ percentage points, and 8 phenomena have a difference less than one point. Moreover, the accuracy difference is larger than the standard deviation of either set of models only on two phenomena: \phen{binding domain} ($+3.2$) and \phen{sentential negation} ($+3.7$). Even for these phenomena, the difference is still less than two standard deviations of at least one of the sets of models.

\begin{table}[h]
\centering
\small
\setlength{\tabcolsep}{4pt}
\begin{tabular}{@{}lr@{\,$\pm$\,}lr@{\,$\pm$\,}lr@{}}
\toprule
Phenomenon & \multicolumn{2}{c}{Ours} & \multicolumn{2}{c}{Published} & Diff. \\
\midrule
Determiner--noun & 94.1 & 0.5 & 93.5 & 0.8 & $+0.5$ \\
Determiner--adj.--noun & 86.3 & 0.4 & 86.2 & 2.1 & $+0.1$ \\
Regular/irregular plural & 92.6 & 0.6 & 92.5 & 0.3 & $+0.2$ \\
Relative-clause distractor & 73.6 & 2.5 & 72.9 & 2.7 & $+0.7$ \\
Relational-noun distractor & 90.1 & 1.7 & 89.7 & 1.3 & $+0.4$ \\
Passive & 86.9 & 0.7 & 86.7 & 0.5 & $+0.2$ \\
Domain & 77.4 & 2.4 & 74.3 & 0.7 & $+3.2$ \\
C-command & 59.1 & 7.9 & 53.3 & 3.3 & $+5.8$ \\
Case & 94.0 & 0.8 & 93.4 & 1.1 & $+0.7$ \\
Reconstruction & 55.4 & 8.7 & 47.2 & 5.8 & $+8.3$ \\
Superlative & 74.6 & 4.8 & 70.1 & 2.9 & $+4.5$ \\
Existential \emph{there} & 96.4 & 1.1 & 95.9 & 1.7 & $+0.5$ \\
Only & 81.9 & 3.2 & 83.0 & 2.7 & $-1.1$ \\
Sentential negation & 76.9 & 2.6 & 73.2 & 2.5 & $+3.7$ \\
Matrix question & 31.0 & 4.2 & 33.3 & 5.1 & $-2.4$ \\
\bottomrule
\end{tabular}
\caption{Held-out BLiMP accuracy (\%) of our reference models and \citet{patil2024filtered}'s published full-corpus models (mean $\pm$ standard deviation over five models). Note that differences are computed between means before rounding occurs.}
\label{tab:local-vs-published}
\end{table}

\section{Phenomena}
\label{app:phenomena}
Table~\ref{tab:phenomena} displays the fifteen phenomena with their BLiMP paradigms, the example pair \citet{patil2024filtered} provide for each paradigm, and each paradigm's number of query and held-out pairs. In addition to the pairs that contain \texttt{<unk>}, we exclude 6 pairs whose two sentences are identical in the released BLiMP data. Table~\ref{tab:direct-share-by-k} (last column) gives the share of the training corpus that each phenomenon's filter marks as direct evidence.
\begin{table*}[p]
\centering
\footnotesize
\setlength{\tabcolsep}{4pt}
\begin{tabularx}{\textwidth}{@{}>{\raggedright\arraybackslash}p{3.0cm}>{\raggedright\arraybackslash\scriptsize}p{3.9cm}>{\raggedright\arraybackslash}X rr@{}}
\toprule
Phenomenon \citep{patil2024filtered} & {\footnotesize BLiMP paradigm} & Example & Query & Held-out \\
\midrule
\multicolumn{5}{@{}l}{\textit{Determiner–noun agreement}} \\
Determiner–noun & \texttt{det\_\allowbreak{}noun\_\allowbreak{}agr\_\allowbreak{}1} & \textit{Craig explored that grocery store/*stores.} & 102 & 706 \\
{\scriptsize\texttt{det-\allowbreak{}noun}} & \texttt{det\_\allowbreak{}noun\_\allowbreak{}agr\_\allowbreak{}2} & \textit{Carl cures those/*that horses.} & 108 & 732 \\
 & \texttt{det\_\allowbreak{}noun\_\allowbreak{}agr\_\allowbreak{}irr\_\allowbreak{}1} & \textit{Phillip was lifting this mouse/*mice.} & 84 & 608 \\
 & \texttt{det\_\allowbreak{}noun\_\allowbreak{}agr\_\allowbreak{}irr\_\allowbreak{}2} & \textit{Those ladies walk through those/*that oases.} & 96 & 665 \\
Determiner–adj.–noun & \texttt{det\_\allowbreak{}noun\_\allowbreak{}agr\_\allowbreak{}with\_\allowbreak{}adj\_\allowbreak{}1} & \textit{Tracy praises those lucky guys/*guy.} & 104 & 725 \\
{\scriptsize\texttt{det-\allowbreak{}adj-\allowbreak{}noun}} & \texttt{det\_\allowbreak{}noun\_\allowbreak{}agr\_\allowbreak{}with\_\allowbreak{}adj\_\allowbreak{}2} & \textit{Some actors buy these/*this gray books.} & 104 & 735 \\
 & \texttt{det\_\allowbreak{}noun\_\allowbreak{}agr\_\allowbreak{}with\_\allowbreak{}adj\_\allowbreak{}irr\_\allowbreak{}1} & \textit{He shouldn't criticize this upset child/*children.} & 88 & 562 \\
 & \texttt{det\_\allowbreak{}noun\_\allowbreak{}agr\_\allowbreak{}with\_\allowbreak{}adj\_\allowbreak{}irr\_\allowbreak{}2} & \textit{That adult has brought that/*those purple octopus.} & 92 & 701 \\
\midrule
\multicolumn{5}{@{}l}{\textit{Subject–verb agreement}} \\
Regular/irregular plural & \texttt{regular\_\allowbreak{}plural\_\allowbreak{}sv\_\allowbreak{}agr\_\allowbreak{}1} & \textit{Jeffrey hasn't/*haven't criticized Donald.} & 100 & 703 \\
{\scriptsize\texttt{agr-\allowbreak{}re-\allowbreak{}irr-\allowbreak{}sv}} & \texttt{regular\_\allowbreak{}plural\_\allowbreak{}sv\_\allowbreak{}agr\_\allowbreak{}2} & \textit{The dress/*dresses crumples.} & 98 & 673 \\
 & \texttt{irr\_\allowbreak{}plural\_\allowbreak{}sv\_\allowbreak{}agr\_\allowbreak{}1} & \textit{This goose isn't/*weren't bothering Edward.} & 85 & 582 \\
 & \texttt{irr\_\allowbreak{}plural\_\allowbreak{}sv\_\allowbreak{}agr\_\allowbreak{}2} & \textit{The woman/*women cleans every public park.} & 109 & 737 \\
Relative-clause distractor\newline{\scriptsize\texttt{agr-\allowbreak{}rel-\allowbreak{}cl}} & \texttt{distractor\_\allowbreak{}agr\_\allowbreak{}relative\_\allowbreak{}clause} & \textit{Boys that aren't disturbing Natalie suffer/*suffers.} & 328 & 337 \\
Relational-noun distractor\newline{\scriptsize\texttt{agr-\allowbreak{}pp-\allowbreak{}mod}} & \texttt{distractor\_\allowbreak{}agr\_\allowbreak{}relational\_\allowbreak{}noun} & \textit{A sketch of lights doesn't/*don't appear.} & 340 & 343 \\
\midrule
\multicolumn{5}{@{}l}{\textit{Argument structure}} \\
Passive & \texttt{passive\_\allowbreak{}1} & \textit{Jeffrey's sons are insulted/*smiled by Tina.} & 150 & 455 \\
{\scriptsize\texttt{passive}} & \texttt{passive\_\allowbreak{}2} & \textit{Most cashiers are disliked/*flirted.} & 157 & 500 \\
\midrule
\multicolumn{5}{@{}l}{\textit{Binding}} \\
Domain & \texttt{principle\_\allowbreak{}A\_\allowbreak{}domain\_\allowbreak{}1} & \textit{Carlos said that Lori helped him/*himself.} & 150 & 729 \\
{\scriptsize\texttt{binding-\allowbreak{}domain}} & \texttt{principle\_\allowbreak{}A\_\allowbreak{}domain\_\allowbreak{}2} & \textit{Mark imagines Erin might admire herself/*himself.} & 138 & 722 \\
 & \texttt{principle\_\allowbreak{}A\_\allowbreak{}domain\_\allowbreak{}3} & \textit{Nancy could say every guy hides himself. / *Every guy could say Nancy hides himself.} & 153 & 751 \\
C-command\newline{\scriptsize\texttt{binding-\allowbreak{}c-\allowbreak{}command}} & \texttt{principle\_\allowbreak{}A\_\allowbreak{}c\_\allowbreak{}command} & \textit{A lot of actresses that thought about Alice healed themselves/*herself.} & 395 & 392 \\
Case & \texttt{principle\_\allowbreak{}A\_\allowbreak{}case\_\allowbreak{}1} & \textit{Tara thinks that she/*herself sounded like Wayne.} & 218 & 633 \\
{\scriptsize\texttt{binding-\allowbreak{}case}} & \texttt{principle\_\allowbreak{}A\_\allowbreak{}case\_\allowbreak{}2} & \textit{Anna imagines herself praising/*praises this boy.} & 210 & 630 \\
Reconstruction\newline{\scriptsize\texttt{binding-\allowbreak{}reconstruction}} & \texttt{principle\_\allowbreak{}A\_\allowbreak{}reconstruction} & \textit{It's herself who Karen criticized / *criticized Karen.} & 462 & 479 \\
\midrule
\multicolumn{5}{@{}l}{\textit{Quantifiers}} \\
Superlative & \texttt{superlative\_\allowbreak{}quantifiers\_\allowbreak{}1} & \textit{No man has revealed more than/*at least 5 forks.} & 213 & 653 \\
{\scriptsize\texttt{quantifier-\allowbreak{}superlative}} & \texttt{superlative\_\allowbreak{}quantifiers\_\allowbreak{}2} & \textit{An/*No actor arrived at at most 6 lakes.} & 220 & 659 \\
Existential there\newline{\scriptsize\texttt{quantifier-\allowbreak{}existential-\allowbreak{}there}} & \texttt{existential\_\allowbreak{}there\_\allowbreak{}quantifiers\_\allowbreak{}1} & \textit{There aren't many/*all lights darkening.} & 414 & 404 \\
\midrule
\multicolumn{5}{@{}l}{\textit{NPI licensing}} \\
Only & \texttt{only\_\allowbreak{}npi\_\allowbreak{}licensor\_\allowbreak{}present} & \textit{Only/*Even Bill would ever complain.} & 191 & 576 \\
{\scriptsize\texttt{npi-\allowbreak{}only}} & \texttt{only\_\allowbreak{}npi\_\allowbreak{}scope} & \textit{Only those doctors who Karla respects ever ... / *Those doctors who only Karla respects ever ...} & 158 & 496 \\
Sentential negation & \texttt{sentential\_\allowbreak{}negation\_\allowbreak{}npi\_\allowbreak{}licensor\_\allowbreak{}present} & \textit{Those banks had not/*really ever lied.} & 202 & 613 \\
{\scriptsize\texttt{npi-\allowbreak{}sent-\allowbreak{}neg}} & \texttt{sentential\_\allowbreak{}negation\_\allowbreak{}npi\_\allowbreak{}scope} & \textit{The turtles that are boring me could not ever ... / *The turtles that are not boring me could ever ...} & 158 & 508 \\
Matrix question\newline{\scriptsize\texttt{npi-\allowbreak{}sim-\allowbreak{}ques}} & \texttt{matrix\_\allowbreak{}question\_\allowbreak{}npi\_\allowbreak{}licensor\_\allowbreak{}present} & \textit{Should I ever join? / *I should ever join.} & 399 & 407 \\
\midrule
Total & & & 5,826 & 18,416 \\
\bottomrule
\end{tabularx}
\caption{The fifteen phenomena, grouped by BLiMP category, with the specific names used in \citet{patil2024filtered}, BLiMP paradigms, and example pairs taken exactly from \citet[Table 1]{patil2024filtered} (* marks the ungrammatical sentence). Paradigm names are abbreviated: agr = agreement, adj = adjective, det = determiner, irr = irregular, sv = subject--verb.}
\label{tab:phenomena}
\end{table*}

\section{Varying Top-$k$}
\label{app:top-k}
Table~\ref{tab:direct-share-by-k} gives the share of direct evidence among each phenomenon's $k$ most influential sentences for other values of $k$.
\begin{table*}[t]
\centering
\small
\begin{tabular}{@{}lrrrrrrrrr@{}}
\toprule
& \multicolumn{8}{c}{Direct share of the top $k$ (\%)} & Whole \\
\cmidrule(lr){2-9}
Phenomenon & 100 & 1k & 3k & 10k & 20k & 40k & 60k & 120k & corpus (\%) \\
\midrule
Determiner--noun & 24.8 & 17.9 & 12.9 & 8.3 & 6.2 & 4.6 & 3.8 & 2.7 & 0.47 \\
Determiner--adj.--noun & 40.0 & 25.9 & 19.0 & 13.4 & 10.4 & 7.7 & 6.4 & 4.5 & 1.15 \\
Regular/irregular plural & 39.2 & 28.6 & 25.6 & 23.0 & 21.4 & 19.7 & 18.7 & 17.0 & 11.62 \\
Relative-clause distractor & \textbf{66.2} & \textbf{54.3} & 45.5 & 32.3 & 24.8 & 18.3 & 15.1 & 10.5 & 2.82 \\
Relational-noun distractor & \textbf{71.0} & \textbf{51.5} & 46.1 & 41.6 & 38.8 & 36.0 & 34.1 & 30.4 & 18.30 \\
Passive & 29.4 & 29.4 & 25.2 & 20.2 & 17.1 & 14.2 & 12.7 & 10.3 & 4.30 \\
Domain & 8.2 & 4.9 & 4.0 & 3.0 & 2.4 & 2.0 & 1.7 & 1.3 & 0.40 \\
C-command & 1.4 & 0.7 & 0.5 & 0.3 & 0.2 & 0.2 & 0.1 & 0.1 & 0.01 \\
Case & 18.0 & 14.8 & 12.0 & 8.5 & 7.2 & 6.0 & 5.4 & 4.3 & 1.41 \\
Reconstruction & $<$0.1 & $<$0.1 & $<$0.1 & $<$0.1 & $<$0.1 & $<$0.1 & $<$0.1 & $<$0.1 & $<$0.01 \\
Superlative & 9.2 & 12.4 & 16.4 & 20.4 & 19.9 & 17.4 & 15.6 & 12.7 & 7.23 \\
Existential \emph{there} & \textbf{90.0} & \textbf{60.9} & 42.1 & 31.1 & 25.3 & 18.7 & 14.9 & 9.3 & 1.17 \\
Only & 11.4 & 6.5 & 4.6 & 2.9 & 2.0 & 1.3 & 1.0 & 0.6 & 0.09 \\
Sentential negation & \textbf{67.8} & 41.5 & 25.0 & 13.5 & 9.8 & 7.0 & 5.7 & 3.6 & 0.52 \\
Matrix question & 12.6 & 6.5 & 4.5 & 2.2 & 1.2 & 0.6 & 0.4 & 0.2 & 0.01 \\
\bottomrule
\end{tabular}
\caption{Share of direct evidence among each phenomenon's $k$ most influential sentences, as a mean over five
reference models. Bold numbers indicate a majority. The last column is the share of all training sentences in the corpus that are marked as direct evidence. This last column figure is used as the baseline number for the enrichment calculation in Figure~\ref{fig:direct-evidence-share}.}
\label{tab:direct-share-by-k}
\end{table*}

\section{Reproducibility, Compute, and AI Usage}
\label{app:compute}

\paragraph{Hardware.} Training for each reference and validation model was performed on a single NVIDIA H100 GPU. EK-FAC attribution was performed with four L40S GPUs per model. Characterization was performed with a mix of H100 GPUs and L40S GPUs. 

\paragraph{Compute.} We trained 275 models for 40 epochs. This includes the five reference models and 270 validation models (150 indirect evidence and matched random models removing 60,000 sentences, 75 BM25 models, 25 direct-evidence models, and 20 models removing 20,000 sentences). One 40-epoch training run took 13 to 27 H100-hours (median 23), and training all 275 models took about 5,700 H100-hours. Superseded and preempted runs add about 270 more H100-hours. EK-FAC fitting and influence functions scoring used 190 L40S-hours. Finally, the characterization analysis used about 8 GPU-hours. Causal validation took the vast majority of compute, while the attribution and characterization steps were relatively cheap. 

\paragraph{Licenses.} The Wikipedia corpus of \citet{gulordava2018colorless} is released under CC BY-NC 4.0, BLiMP of \citet{warstadt2020blimp} under CC BY 4.0, and the Filtered Corpus Training code of \citet{patil2024filtered} under GPL-3.0. Our usage of these resources for non-commercial academic research complies with these licenses.

\paragraph{Software.} We train our language models with PyTorch 2.5.1 and Transformers 4.46.3. We perform attribution using EK-FAC influence with Kronfluence 1.0.1. We perform parsing with Stanza 1.13. We score BM25 with rank-bm25 0.2.2.

\paragraph{AI usage.} We used Claude and ChatGPT as AI assistants for coding and writing. AI assisted with coding under the authors' direction and supervision. Relatedly, AI was used to critique the authors' proposed analyses. The authors had the final decision on which analyses to run. AI was used for editing and proofreading of the paper during drafting and revision, for formatting, and for producing TikZ diagrams. Similarly, AI was also used to help while drafting low-novelty sections of the appendix (compute, experimental details, controls) using our code, configurations, and run records. The authors take full responsibility for the content of this paper and its code. 

\section{Direct-Evidence Filters}
\label{app:filters}

We reproduce the twelve parse-based filters used in \citet{patil2024filtered} following their codebase. We also reproduce their three NPI surface filters as described in \citet{patil2024filtered}. Each parse-based filter utilizes a sentence's Stanza dependency parse \citep{qi2020stanza}. These parse-based filters perform filtering based on dependency relations, morphological features, and, for some phenomena, lexical lists. One slight deviation is that we parse with Stanza 1.13 and remove each sentence's end-of-sentence token, whereas \citet{patil2024filtered} used Stanza 1.5 and kept the end-of-sentence token.

For computing the structural coverage metric in Figure~\ref{fig:structural-coverage}, we split the direct evidence filter's logic into necessary conditions that must all be true for the filter to hold, and when the original logic has a disjunction, we have alternative branching conditions where at least one of these branches should be satisfied for direct evidence. We list these filter conditions in Table~\ref{tab:filter-conditions}. Our filters mark direct-evidence shares of the training corpus within 0.33 percentage points of the figures reported in \citet[Table~1]{patil2024filtered} for every phenomenon except \phen{passive} (4.30\% vs.\ 2.67\%). This difference for the passive construction may be caused by the updated Stanza version's parsing differences or by our removal of the end-of-sentence token.

\paragraph{Structural coverage.} Figure~\ref{fig:structural-coverage} omits four phenomena due to their filters structurally leaving little to no variance in coverage, meaning that comparison would be uninformative. For completeness of results, we provide results for these four phenomena in Table~\ref{tab:omitted-structural-coverage}. For \phen{relative-clause distractor}, \phen{relational-noun distractor} and \phen{reconstruction}, the reason for zero variance is that any sentence that has the structural condition is direct evidence. Therefore, for these three phenomena, all indirect evidence must have structural coverage of $0$. For \phen{regular/irregular plural}, the only structural condition in the filter is having a subject, which nearly every sentence in the corpus has, so every group's structural coverage is near $1$, making this phenomenon also uninformative for comparison.

\begin{table}[h]
\centering
\small
\begin{tabular}{@{}lccc@{}}
\toprule
Phenomenon & Indirect & BM25 & \begin{tabular}[b]{@{}c@{}}Matched\\random\end{tabular} \\
\midrule
Regular/irregular plural   & 0.988 & 0.990 & 0.962 \\
Relative-clause distractor & 0     & 0     & 0     \\
Relational-noun distractor & 0     & 0     & 0     \\
Reconstruction             & 0     & 0     & 0     \\
\bottomrule
\end{tabular}
\caption{Structural coverage for the four phenomena omitted from
Figure~\ref{fig:structural-coverage}, as means computed over 5 reference models.}
\label{tab:omitted-structural-coverage}
\end{table}

\begin{table*}[t]
\centering
\footnotesize
\setlength{\tabcolsep}{4pt}
\begin{tabularx}{\textwidth}{@{}>{\raggedright\arraybackslash}p{3.0cm}c>{\raggedright\arraybackslash}X>{\raggedright\arraybackslash}X r@{}}
\toprule
Phenomenon & & Structural conditions (counted) & Lexical conditions (not counted) & Max. \\
\midrule
\multicolumn{5}{@{}l}{\textit{Determiner–noun agreement}} \\
Determiner–noun &  & demonstrative determiner (\textit{this, that, these, those}) & its noun is on the BLiMP noun list (450 words) & 1 \\
Determiner–adj.–noun &  & demonstrative determiner (\textit{this, that, these, those}); the demonstrative is directly followed by a non-noun (e.g., an adjective) & -- & 0.5 \\
\midrule
\multicolumn{5}{@{}l}{\textit{Subject–verb agreement}} \\
Regular/irregular plural &  & the sentence has a subject & a subject is on the BLiMP subject list (729 words, including names) & 1 \\
Relative-clause distractor &  & a relative clause modifies the subject or a noun modifying it & -- & 0 \\
Relational-noun distractor &  & a prepositional modifier of the subject stands between the subject and its predicate & -- & 0 \\
\midrule
\multicolumn{5}{@{}l}{\textit{Argument structure}} \\
Passive & (a) & a passive verb & the verb is on the BLiMP passive list (127 verbs) & 1 \\
 & (b) & a copula with a predicate (e.g., \textit{are disliked} parsed as an adjective) & the predicate is on the same list &  \\
\midrule
\multicolumn{5}{@{}l}{\textit{Binding}} \\
Domain & (a) & a complementizer (e.g., \textit{that}) introducing a complement clause; an object or oblique of the clause's head, after it & the object is a pronoun & 1 \\
 & (b) & a reflexive depending on the head of a complement clause (\texttt{ccomp} or \texttt{xcomp}) & a reflexive (\textit{-self}) &  \\
C-command &  & a reflexive whose head also has a subject; a relative clause inside that subject, before the reflexive & a reflexive (\textit{-self}) & 0.5 \\
Case & (a) & a complementizer (e.g., \textit{that}) introducing a complement clause; the clause's subject directly follows the complementizer & the subject is a pronoun & 1 \\
 & (b) & a reflexive directly followed by a clause attached to the reflexive's head (\texttt{xcomp} or \texttt{advcl}) & a reflexive (\textit{-self}) &  \\
Reconstruction &  & a reflexive heading a clause with a subject (\textit{It's herself who \dots}) & a reflexive (\textit{-self}) & 0 \\
\midrule
\multicolumn{5}{@{}l}{\textit{Quantifiers}} \\
Superlative &  & a comparative or superlative word (e.g., \textit{more}, \textit{fewer}); that word is inside an object or oblique & -- & 0.5 \\
Existential there &  & existential \textit{there} with \textit{be}; that \textit{be} has a subject with a determiner or modifier & the subject's quantifier is one of BLiMP's ten (\textit{a, an, no, some, few, many, all, most, every, each}) & 1 \\
\midrule
\multicolumn{5}{@{}l}{\textit{NPI licensing}} \\
Only &  & \textit{only} occurs & an NPI from the list follows it & 1 \\
Sentential negation &  & a negation (\textit{not}, \textit{cannot}, \textit{n't}) & an NPI from the list follows it & 1 \\
Matrix question &  & the sentence ends in a question mark & the sentence contains an NPI from the list & 1 \\
\bottomrule
\end{tabularx}
\caption{Conditions for each direct-evidence filter. A sentence is considered direct evidence by the filter if it meets all conditions in at least one branch (a, b). The structural coverage of a sentence is the fraction of a branch's structural conditions it meets, taking the best branch. Lexical conditions (BLiMP and NPI word lists, and whether a word is a pronoun or a reflexive) do not count towards structural coverage. Max.\ is the highest coverage of any sentence outside the direct-evidence filter, and thus indicates an upper limit for the structural coverage metric. We reproduce the parse-based filters from \citet{patil2024filtered} and split them into conditions. The three NPI filters used are surface filters following \citet{patil2024filtered}, and their structural coverage is the presence of the licensing cue.}
\label{tab:filter-conditions}
\end{table*}

\paragraph{NPI filters.} We elaborate more on the construction of the surface filters for the NPIs. \citet[\S3.2.1]{patil2024filtered} describe the three NPI filters: \texttt{npi-only} removes sentences with an NPI after \textit{only}, \texttt{npi-sent-neg} sentences with a negation and an NPI, and \texttt{npi-sim-ques} questions with NPIs. As they do, we take NPIs from \citet{jumelet2021monotonicity}, which are the 40 expressions in its Appendix~A (e.g., \textit{any}, \textit{ever}, \textit{yet}). The filter checks for matches in the lowercased sentence with a space on each side. Following that paper's filtering rule, a sentence is direct evidence when \textit{only} or a negation (\textit{not}, \textit{cannot}, \textit{n't}) occurs anywhere before an NPI or, for \phen{matrix question}, when it contains an NPI and ends in a question mark.
\paragraph{Examples of indirect evidence.} We elaborate more on the examples of indirect evidence provided in Table~\ref{tab:structural-similarity-examples}. In general, some indirect evidence sentences are the phenomenon itself with words outside the filter's lists: \textit{those products} is \phen{determiner--noun agreement}, but \textit{products} is not on the BLiMP noun list. This is because of the lexical generalization design of \citet{patil2024filtered}. Other indirect evidence sentences are related constructions: in \textit{Several of ESPN's personalities also portray themselves}, the reflexive agrees with the subject, like the \phen{c-command} pairs, but the relevant noun is in a prepositional phrase rather than a relative clause. 
\section{Control groups}
\label{app:groups}\label{app:matching}

\paragraph{Matched random.} We match indirect evidence sentences by their length in tokens and its mean word frequency: the average over its tokens of $\log\frac{c(w)+1}{N+V}$, the add-one smoothed log probability of token $w$, where $c(w)$ is the token $w$'s count in the training corpus, $N$ the corpus size in tokens, and $V$ the vocabulary size. Possible candidates for the matched random control group are the non-direct sentences outside the indirect evidence. Candidates are selected without replacement and each is used at most once in the matched random control group. For each indirect evidence sentence, we randomly pick uniformly one of the 10 candidates of the same length with the closest mean frequency. If no candidate of that exact length remains in the pool, we choose one of the 10 training sentences which are nearest on both covariates. In the resulting matched random control group, sentence lengths are matched almost exactly, and the standardized mean difference in mean frequency is about 0.02 (0.04 for the 20,000-sentence groups of Appendix~\ref{app:removal-size}).

\paragraph{BM25.} We compute BM25 scores for each training sentence with BM25Okapi from rank-bm25 0.2.2 ($k_1=1.5$, $b=0.75$, $\epsilon=0.25$). We split sentences on whitespace. We consider a phenomenon's query to be the set of words in its query pairs, from both of the sentences in a minimal pair. The BM25 group is the 60,000 highest-scoring non-direct sentences. Note that the BM25 control group does not depend on the reference model and is shared between reference models; the trained BM25 models differ because of different initializations. The lexical similarity in Figure~\ref{fig:lexical-similarity} is a group's mean BM25 score. Thus, by definition, the BM25 group has the highest lexical similarity of any 60,000 sentences out of the pool of non-direct sentences.

\paragraph{Direct evidence.} For retraining, the direct-evidence group is a uniform random sample of 60,000 sentences that match the direct-evidence filter. This control group does not depend on the reference model and is shared between reference models. There are five phenomena with at least 60,000 direct-evidence sentences: \phen{relational-noun distractor}, \phen{relative-clause distractor}, \phen{regular/irregular plural}, \phen{superlative}, and \phen{passive}. Importantly, we are comparing removals of equal size rather than removing all direct evidence as was done in \citet{patil2024filtered}; the direct evidence control group removes between 13\% (\phen{relational-noun distractor}) and 87\% (\phen{relative-clause distractor}) of their direct evidence. Our direct evidence retraining results should not be taken to be comparable to \citet{patil2024filtered}'s results.

\section{Additional validation results}
\label{app:validation}

\paragraph{Results by phenomenon.} Table~\ref{tab:validation} gives the held-out accuracies behind Figure~\ref{fig:removal-accuracy}.

\begin{table*}[t]
\centering
\small
\begin{tabular}{@{}lrrrrrr@{\,$\pm$\,}l@{}}
\toprule
& & \multicolumn{4}{c}{After removing} & \multicolumn{2}{c}{} \\
\cmidrule(lr){3-6}
Phenomenon & Reference & Matched random & Indirect & BM25 & Direct & \multicolumn{2}{c}{Extra drop} \\
\midrule
Determiner--noun & 94.1 $\pm$ 0.5 & 94.2 $\pm$ 0.5 & 85.6 $\pm$ 1.1 & 94.2 $\pm$ 0.7 & -- & 8.6 & 0.6 \\
Determiner--adj.--noun & 86.3 $\pm$ 0.4 & 86.0 $\pm$ 1.5 & 78.3 $\pm$ 1.5 & 85.6 $\pm$ 0.9 & -- & 7.7 & 1.6 \\
Regular/irregular plural & 92.6 $\pm$ 0.6 & 93.1 $\pm$ 0.5 & 89.9 $\pm$ 0.6 & 93.0 $\pm$ 0.5 & 92.3 $\pm$ 0.6 & 3.2 & 0.9 \\
Relative-clause distractor & 73.6 $\pm$ 2.5 & 76.5 $\pm$ 1.4 & 64.2 $\pm$ 1.9 & 78.2 $\pm$ 2.4 & 69.1 $\pm$ 3.4 & 12.3 & 2.3 \\
Relational-noun distractor & 90.1 $\pm$ 1.7 & 90.1 $\pm$ 2.1 & 80.4 $\pm$ 2.6 & 90.0 $\pm$ 1.5 & 90.0 $\pm$ 1.9 & 9.7 & 3.2 \\
Passive & 86.9 $\pm$ 0.7 & 86.6 $\pm$ 1.3 & 80.8 $\pm$ 1.6 & 86.5 $\pm$ 1.1 & 84.0 $\pm$ 1.2 & 5.8 & 2.7 \\
Domain & 77.4 $\pm$ 2.4 & 77.9 $\pm$ 1.8 & 61.7 $\pm$ 1.1 & 77.3 $\pm$ 1.5 & -- & 16.2 & 1.6 \\
C-command & 59.1 $\pm$ 7.9 & 62.1 $\pm$ 5.3 & 33.6 $\pm$ 0.7 & 60.8 $\pm$ 7.2 & -- & 28.6 & 5.2 \\
Case & 94.0 $\pm$ 0.8 & 94.4 $\pm$ 0.9 & 82.7 $\pm$ 1.7 & 92.7 $\pm$ 0.8 & -- & 11.7 & 2.1 \\
Reconstruction & 55.4 $\pm$ 8.7 & 50.7 $\pm$ 4.3 & 6.3 $\pm$ 2.2 & 58.7 $\pm$ 7.2 & -- & 44.4 & 5.7 \\
Superlative & 74.6 $\pm$ 4.8 & 72.5 $\pm$ 4.6 & 40.1 $\pm$ 8.2 & 74.3 $\pm$ 6.5 & 71.3 $\pm$ 6.1 & 32.4 & 8.7 \\
Existential \emph{there} & 96.4 $\pm$ 1.1 & 96.8 $\pm$ 1.1 & 92.4 $\pm$ 1.3 & 96.7 $\pm$ 1.4 & -- & 4.4 & 1.9 \\
Only & 81.9 $\pm$ 3.2 & 83.9 $\pm$ 4.3 & 42.0 $\pm$ 6.9 & 82.4 $\pm$ 3.9 & -- & 41.9 & 4.6 \\
Sentential negation & 76.9 $\pm$ 2.6 & 74.8 $\pm$ 4.6 & 57.4 $\pm$ 3.6 & 73.5 $\pm$ 5.1 & -- & 17.4 & 6.8 \\
Matrix question & 31.0 $\pm$ 4.2 & 33.0 $\pm$ 4.3 & 3.7 $\pm$ 1.8 & 22.8 $\pm$ 4.2 & -- & 29.3 & 5.3 \\
\bottomrule
\end{tabular}
\caption{Held-out BLiMP accuracy (\%) of the reference models and of the models retrained without each group, as mean $\pm$ standard deviation over the five models. Direct evidence is a random 60,000 of its sentences, removed only for the five phenomena with at least that many. Extra drop is the accuracy lost by removing the indirect evidence relative to the matched random group: mean $\pm$ standard deviation of the paired difference over the five models.}
\label{tab:validation}
\end{table*}

\paragraph{Significance test.}  Our reported statistical test in Section~\ref{sec:validation} is an exact one-sided sign-flip test over the five reference models. The sign-flip test is on held-out loss, the logistic margin loss of \S\ref{sec:objective}. For each reference model, we average over the 15 phenomena the extra loss from removing the indirect evidence relative to removing the matched random group. We compare the observed mean paired difference with all $2^5 = 32$ sign assignments. Thus, the smallest possible $p$ from this sign-flip test is $1/32$, which we achieve.

\paragraph{Specificity.} Figure~\ref{fig:specificity} shows how much each phenomenon's held-out accuracy drops when any phenomenon's indirect evidence is removed. The diagonal gives each phenomenon's own drop (median 11.4 points). The median of off-diagonal cells is 0.3. The three largest off-diagonal drops are within categories. For example, \phen{determiner--noun} and \phen{determiner--adjective--noun agreement} hurt each other by 8.3 and 8.2 points (the two differ structurally only by the intervening adjective, and their indirect evidence overlaps by $38\%$). 

\paragraph{Checkpoint.}
\label{app:par:checkpoint}
In Section~\ref{sec:validation}, we evaluated each validation model at its reference model's selected epoch. One may ask if this choice of when to evaluate the validation model could materially impact the results. We find that it does not change our conclusions. Our results do not change much if we evaluate the validation model instead at its own lowest-validation-loss epoch. Removing the indirect evidence lowers accuracy more than the matched random group in all 75 model pairs. Averaged over the five initializations, the range of the extra drop is 3.3 to 45.6 points (median 12.0). Similarly, our results also do not change much if we evaluate after all $40$ epochs. When evaluating after $40$ epochs, removing the indirect evidence lowers accuracy more than the matched random group in all 75 model pairs. Averaged over the five initializations, the range of the extra drop is 3.2 to 44.1 points (median 12.4). Thus, this suggests our findings are robust to changing to various choices of when we evaluate the validation models.

\paragraph{Removal size.}\label{app:removal-size} We want to understand the effect of varying the hyperparameter $k$, which is the size of the indirect evidence set that we study. For two phenomena, we also removed the 20,000 most influential non-direct sentences and a matched random group of the same size. Table~\ref{tab:removal-size} shows a smaller but still positive extra drop in accuracy and extra loss compared to a matched random group of the same size. While we only performed this comparison for two phenomena, this result does suggest that our methodology should find causally important indirect evidence sets at various $k$. 

\begin{table}[h]
\centering
\small
\begin{tabular}{@{}llcc@{}}
\toprule
& & \multicolumn{2}{c}{Extra harm} \\
\cmidrule(l){3-4}
Phenomenon & Removed & Loss & Acc.\ (pts) \\
\midrule
Determiner--noun & 20,000 & $+0.130$ & 5.4 \\
                 & 60,000 & $+0.217$ & 8.6 \\
Regular/irregular plural & 20,000 & $+0.041$ & 1.9 \\
                         & 60,000 & $+0.102$ & 3.2 \\
\bottomrule
\end{tabular}
\caption{A comparison of the effect of removing the most influential indirect evidence at two different sizes, as means over five models. The displayed extra drop for held-out loss and held-out accuracy are relative to the matched random control group of the same size.}
\label{tab:removal-size}
\end{table}

\section{Shared indirect evidence}
\label{app:sharing}
\begin{figure}[!t]
    \centering
    \includegraphics[width=\columnwidth]{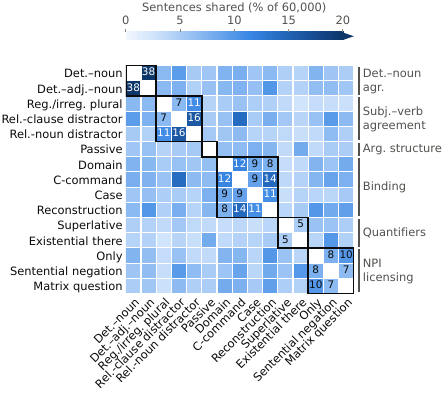}
    \caption{Overlap between phenomena's indirect evidence. Each cell in the heatmap is the share of one phenomenon's 60,000 indirect evidence sentences that are also in the other's indirect evidence, averaged over five reference models. Black outlines mark BLiMP categories.}
    \label{fig:category-membership}
\end{figure}

Figure~\ref{fig:category-membership} shows the overlap between the indirect evidence of every pair of phenomena. The clustering in Figure~\ref{fig:category_dendrogram} uses this same overlap metric.  

\section{Influence estimation details}
\label{app:influence}
Table~\ref{tab:ekfac-settings} lists the EK-FAC settings, shared by all five reference models.

\begin{table}[!h]
\centering
\small
\begin{tabularx}{\columnwidth}{@{}l>{\raggedright\arraybackslash}X@{}}
\toprule
Setting & Value \\
\midrule
Curvature & EK-FAC, empirical Fisher \\
Tracked layers & Query, key, value and output projections, both feed-forward layers in the 8 blocks \\
Not tracked & Token and position embeddings, layer norms, output layer  \\
Factor fitting & All 2,442,181 training sentences \\
Damping & $0.1\times$ the mean eigenvalue of each layer \\
Loss for training gradient $g_i$ & Summed next-token cross-entropy of training sentence $i$ \\
Query gradient $q$ & Gradient of $\mathcal{L}_t$ (\S\ref{sec:objective}) over the query set \\
Checkpoint & Lowest validation loss (epochs 14, 14, 14, 15, 17) \\
Hardware & Four NVIDIA L40S GPUs per reference model \\
\bottomrule
\end{tabularx}
\caption{EK-FAC settings, shared by all five reference models.}
\label{tab:ekfac-settings}
\end{table}

\paragraph{Fisher matrix.} The EK-FAC method uses the Fisher matrix to approximate the curvature term in the efficient computation of influence functions scores. There are two methods of calculating the Fisher matrix: the empirical Fisher and the true Fisher. In this study, we use the empirical Fisher which uses gradients for the observed next tokens in the corpus, while the true Fisher uses gradients for next tokens sampled from the model's prediction distribution. The two methods of calculating the Fisher matrix are the same in expectation when the model's prediction distributions match the data distribution. In a preliminary comparison on a published model from \citet{patil2024filtered}, we scored \phen{determiner--noun agreement} against a 500,000-sentence sample of the corpus with both variants and fit the factors on 8,192 sentences. We obtained almost identical rankings (Spearman $\rho=0.998$), with the top 60,000 influential sentences having an overlap of $97\%$. This result indicates that our choice of how to calculate the Fisher matrix does not materially impact our study; we only depend on the ranking of the training sentences obtained from influence functions scores, and we do not directly utilize the magnitude of influence functions scores for our analysis.

\end{document}